\documentclass[letterpaper, 10 pt, journal, twoside]{ieeetran} 

\IEEEoverridecommandlockouts                              %

\usepackage{amsmath}

\usepackage{tikz}
\usetikzlibrary{shapes,arrows, calc, arrows.meta, intersections, positioning, patterns, decorations.pathreplacing, backgrounds} %
\usepackage{selectp} %
\usepackage{mathtools}
\usepackage{subcaption}
\usepackage{amsfonts}
\usepackage[noend]{algorithm2e}
\usepackage{calculator}
\usepackage{siunitx}
\usepackage{stfloats}
\usepackage{soul}  %
\usepackage[absolute,overlay]{textpos}
\usepackage{cite}
\usepackage{makecell, multirow}
\usepackage{comment}
\setcellgapes{1.1pt}
\usepackage{booktabs}
\usepackage{soul}
\newcommand{\tabitem}{~~\llap{\textbullet}~~}

\newcommand*\rot{\rotatebox{90}}

\definecolor{LINE_COLOR_RED}{RGB}{214,39,40}
\definecolor{LINE_COLOR_BLUE}{RGB}{0,0,150}
\definecolor{LINE_COLOR_GREEN}{RGB}{44,160,44}

\definecolor{TABLE_ENTRY_ONE}{RGB}{214,39,40}
\definecolor{TABLE_ENTRY_TWO}{RGB}{0,140,0}
\definecolor{TABLE_ENTRY_THREE}{RGB}{0,0,150}
\definecolor{TABLE_ENTRY_FOUR}{RGB}{139,69,13} %

\definecolor{TABLE_ENTRY_STATE_ACTION_RISK}{RGB}{0,0,150}
\definecolor{TABLE_ENTRY_STATE_RISK}{RGB}{0,140,0}

\def\linkToCode{github.com/translearn/safeMotionsRisk}
\def\linkToVideo{https://youtu.be/fR9SByhCbDI}
\RestyleAlgo{ruled}

\usepackage{hyperref}

\title{
Safe Reinforcement Learning of Robot Trajectories\\in the Presence of Moving Obstacles
}

\author{Jonas Kiemel$^{1}$, Ludovic Righetti$^{2}$, Torsten Kröger and Tamim Asfour$^{1}$%
\thanks{Manuscript received: June, 10, 2024; Revised September, 15, 2024; Accepted October, 5, 2024.}%
\thanks{This paper was recommended for publication by Editor Jaydev P. Desai upon evaluation of the Associate Editor and Reviewers' comments.
This work was supported by a research travel grant of the DAAD-Stiftung.} %
\thanks{$^{1}$Institute for Anthropomatics and Robotics (IAR), Karlsruhe Institute of Technology (KIT), Germany,
        {\tt\footnotesize jonas.kiemel@kit.edu}}%
\thanks{$^{2}$Tandon School of Engineering, New York University (NYU), USA
        }%
\thanks{Accompanying video: \href{\linkToVideo}{\linkToVideo}}
}

\begin{document}

\maketitle
\markboth{IEEE Robotics and Automation Letters. Preprint Version. Accepted October, 2024}
{Kiemel \MakeLowercase{\textit{et al.}}: Safe Reinforcement Learning of Robot Trajectories in the Presence of Moving Obstacles}  

\begin{textblock*}{14.0cm}(3.4cm,0.3cm) 
	{\tiny © 2024 IEEE.  Personal use of this material is permitted.  Permission from IEEE must be obtained for all other uses, in any current or future media, including reprinting/republishing this \\[-1.0em] material for advertising or promotional purposes, creating new collective works, for resale or redistribution to servers or lists, or reuse of any copyrighted component of this work in other works.}
\end{textblock*}

\vspace*{-0.74cm}
\begin{abstract}
In this paper, we present an approach for learning collision-free robot trajectories in the presence of moving obstacles. 
As a first step, we train a backup policy to generate evasive movements from arbitrary initial robot states using model-free reinforcement learning. 
When learning policies for other tasks, the backup policy can be used to estimate the potential risk of a collision and to offer an alternative action if the estimated risk is considered too high.
No matter which action is selected, our action space ensures that the kinematic limits of the robot joints are not violated. 
We analyze and evaluate two different methods for estimating the risk of a collision.
A physics simulation performed in the background is computationally expensive but provides the best results in deterministic environments. 
If a data-based risk estimator is used instead, the computational effort is significantly reduced, but an additional source of error is introduced.
For evaluation, we successfully learn a reaching task and a basketball task while keeping the risk of collisions low. 
The results demonstrate the effectiveness of our approach for deterministic and stochastic environments, including a human-robot scenario and a ball environment, where no state can be considered permanently safe. 
By conducting experiments with a real robot, we show that our approach can generate safe trajectories in \mbox{real time}. %
\end{abstract}

\begin{IEEEkeywords}
Motion Control, Reinforcement Learning, Robot Safety, Collision Avoidance
\end{IEEEkeywords}
\section{INTRODUCTION}
\IEEEPARstart{I}{n} recent years, model-free reinforcement learning (RL) has become increasingly popular for generating robot trajectories in real time.
This trend is mainly driven by the fact that model-free RL can be easily applied to a wide range of robotic applications.
Instead of using a differentiable model of the system dynamics, well-performing actions are identified during a training phase based on trial and error. 
When performing movements with a real robot, however, it is important to ensure that the selected actions do not cause damage. %
One approach to avoid safety violations is to execute an action only if the robot will remain in a safe state after executing the action. %
However, when a robot is operating in an environment with moving obstacles, determining whether the subsequent state is safe becomes non-trivial: 
The exact motion of the obstacles might not always be known in advance and the kinematic constraints of the robot joints may limit its ability to execute arbitrary evasive movements.
In modern model-free RL, actions for a desired learning task are typically generated by a task policy, which is parameterized by a neural network. 
To prevent collisions during and after the training phase of the task policy, we propose to additionally use a backup policy trained on avoiding obstacles. %
As shown in Fig.~\ref{fig:three_steps}, the backup policy serves two purposes.
First, an action from the backup policy is executed if the task policy would lead to an unsafe follow-up state.
Second, a rollout of the backup policy is used to assess whether the follow-up state is safe.
While the rollout can be carried out in a physics simulator, we additionally evaluate the performance of neural networks trained to estimate collision risks based on data from previous rollouts. 
This becomes important when dealing with stochastic environments and reduces the computational effort required for real-time execution.
\begin{figure}[t]
    \vspace{-0.45cm}
    \input{figures/header/header}
    \vspace{-0.4cm}
	\caption{%
	Our approach shown for the \textit{Ball} environment, where balls are thrown towards the robot from random directions.} %
    
	\label{fig:three_steps}
 \vspace{-0.5cm}
\end{figure}
The main contributions of the work are: 
\begin{itemize}
\item 
We propose an approach for learning safe goal-directed motions in environments with moving obstacles while considering the kinematic constraints of robot joints. 

\item 
We investigate the impact of stochastically moving obstacles, compare different methods to estimate collision risks, analyze potential reasons for incorrect estimates, and discuss measures to prevent them. 
\item 
We conduct a systematic quantitative analysis of the presented approach by learning a reaching task and a \mbox{basketball} task in three different environments. %
Experiments with a real robot demonstrate that safe trajectories can be generated in real time. 
\end{itemize} 

\noindent Our code is available  \mbox{at {\href{https://www.\linkToCode}{\linkToCode}}}.

\begin{table*}[t]
    \vspace{-0.15cm}
    \caption{Overview of related approaches that address instantaneous safety constraints by adjusting actions of an RL agent.}
    \vspace{-0.15cm}
    \makegapedcells
\begin{tabular*}{\textwidth}{@{}p{28.0mm}p{46.5mm}p{46mm}p{44mm}} 
    \toprule

     & \multicolumn{1}{c}{Action adjustment using}  & \multicolumn{1}{c}{Application scenario}  & \multicolumn{1}{c}{Main difference to our approach}  %
     \\ \hline
\hspace{-0.05cm}\tabitem Pham et al.~\cite{pham2018optlayer} & Faverjon and Tournassoud’s method~\cite{faverjon1987local} & Reaching task with an industrial robot & Problem of conflicting constraints \\
\hspace{-0.05cm}\tabitem Fisac et al.~\cite{fisac2018general} & Reachability analysis & Quadrotor control & Model-based safety controller\\
\hspace{-0.05cm}\tabitem Wabersich et al.~\cite{wabersich2021predictive} & Model-predictive control & Pendulum swing-up, quadrotor control & Model-based safety filter\\
\hspace{-0.05cm}\tabitem Kiemel et al.~\cite{kiemel2022learning} & Braking trajectories & Reaching task with an industrial robot & No consideration of moving obstacles \\
\hspace{-0.05cm}\tabitem Yang et al.~\cite{yang2022safe} & Backup policy trained via model-free RL & Locomotion with a quadruped robot & Model-based risk criteria\\
\hspace{-0.05cm}\tabitem Thananjeyan et al.~\cite{thananjeyan2021recovery}$\!\!\!$ & Backup policy trained via model-free RL & Navigation and manipulation tasks  & Backup policy depends on task policy \\
    \bottomrule
    \end{tabular*}
\label{table:related_work}
\vspace{-0.35cm}
\end{table*}

\section{Related work}

The recent success of model-free RL in simulation environments has sparked a growing interest in research addressing the challenges of safe reinforcement learning with real robots. 
In model-free RL, safety constraints are typically formalized using a constrained Markov decision process (CMDP)~\cite{altman1999constrained}.
In this context, a distinction is made between cumulative constraints~\cite{achiam2017constrained, liu2020ipo}, which are defined with respect to a penalty received over time, and instantaneous constraints, which must be satisfied at each decision step.
A survey of various approaches to increase safety during and after the training phase can be found in~\cite{brunke2022safe}. 
This work focuses on addressing instantaneous constraints by adjusting risky actions of an RL agent. 
An overview of existing approaches in this field and their differences to the approach presented in this work is provided in TABLE~\ref{table:related_work}.

Pham et al.~\cite{pham2018optlayer} utilize Faverjon and Tournassoud’s method~\cite{faverjon1987local} to avoid collisions with moving obstacles by solving a quadratic program~(QP). 
This approach, however, does not ensure recursive feasibility, meaning that the QP may not have a solution due to conflicting constraints.
For the specific case of collision avoidance in a two-dimensional plane, Zhao et al.~\cite{zhao2021model} addressed the problem of recursive feasibility using barrier certificates. 
More generally, techniques from model-based control can be used to increase safety when using RL~\cite{berkenkamp2017safe, koller2018learning, cheng2019end}.
For example, Fisac et al.~\cite{fisac2018general} utilize Hamilton–Jacobi reachability analysis to avoid ground contact with a quadrotor.  
Wang et al.~\cite{wang2023safe} use barrier functions to avoid collisions with static obstacles in the context of real-time navigation.
Wabersich et al.~\cite{wabersich2021predictive} utilize a safety filter based on model-predictive control to swing-up an inverted pendulum and to control a quadrotor in a simulation environment. 
The safety filter modifies actions if no safely executable backup trajectory with a specified time horizon is found otherwise.
While our approach is also based on the idea of safe backup trajectories, we utilize a backup policy trained via model-free RL to adjust risky actions.
The use of a model-free backup policy offers the advantage that no differentiable model of the system dynamics needs to be specified~\cite{hans2008safe}.
Moreover, generating an action with a backup policy represented by a neural network requires little computational effort. 
In one of our previous works~\cite{kiemel2022learning}, we utilized time-optimal braking trajectories as backup trajectories~\cite{ShortPaperMotionSafety, rubrecht2012motion}. 
The approach reliably prevents self-collisions and collisions with static obstacles, but does not account for moving obstacles. 
To overcome this limitation, we utilize model-free RL to train a backup policy that actively avoids collisions with moving obstacles which may behave stochastically. 
The use of model-free RL is related to Yang et al.~\cite{yang2022safe}, where a backup policy is trained to stabilize a quadruped robot, and to Thananjeyan et al.~\cite{thananjeyan2021recovery}, where backup policies are used for navigation and manipulation tasks. 
In contrast to Yang et al.~\cite{yang2022safe}, our approach does not make use of model-based risk criteria. Instead, the risk of an action is determined by performing a background simulation or by utilizing a data-based risk estimator~\cite{srinivasan2020learning}.
Compared to Thananjeyan et al.~\cite{thananjeyan2021recovery}, our backup policy does not depend on the current task policy. 
As a result, the backup policy can be used to learn different tasks without needing to be updated. 
By utilizing a special action space for both the task policy and the backup policy~\cite{kiemel2021learning}, we additionally ensure that the resulting trajectories are jerk-limited no matter which policy is selected. 
In our evaluation section, we provide comparative results with two alternative approaches~\cite{pham2018optlayer, kiemel2022learning} that have been used in the context of industrial robotics.

\section{Problem statement}
We assume that model-free RL is used to learn motions for a robotic manipulator. 
During and after the training phase, self-collisions and collisions with static or moving obstacles must be avoided. 
In addition, the following kinematic constraints must be satisfied  by each robot joint at all times:
\begin{alignat}{3}
\mathrm{p}_{\mathrm{min}} &{}\le{}& \mathrm{\theta} &{}\le{}& \mathrm{p}_{\mathrm{max}}\phantom{,}    \label{eq:constraint_p} \\ 
\mathrm{v}_{\mathrm{min}} &{}\le{}& \dot{\mathrm{\theta}} &{}\le{}& \mathrm{v}_{\mathrm{max}}\phantom{,}    \label{eq:constraint_v}\\
\mathrm{a}_{\mathrm{min}} &{}\le{}& \ddot{\mathrm{\theta}}&{}\le{}& \mathrm{a}_{\mathrm{max}}\phantom{,}  \label{eq:constraint_a}  \\
\mathrm{j}_{\mathrm{min}} &{}\le{}& \dddot{\mathrm{\theta}} &{}\le{}& \mathrm{j}_{\mathrm{max}},  \label{eq:constraint_j}
\end{alignat}
where $\theta$ is the joint position. 

As common for industrial robots, the base of the manipulator is assumed to be fixed. 
In the most general case, a motion determined by an action of an RL agent can be safely executed if an infinite \mbox{follow-up} trajectory is known, such that the composed movement satisfies the desired safety constraints \cite{ShortPaperMotionSafety}. 
If no such trajectory exists, the action of the RL agent needs to be adjusted.
While reasoning over infinite trajectories is impractical, the condition can be relaxed if safe goal states exist. In this context, a safe goal state is a state in which a safe \mbox{follow-up} trajectory is known.
The most trivial case is a resting state, provided that the robot is stopped in a region of the environment where no moving obstacles are present. 
The better a method performs in finding safe follow-up trajectories, the less actions of the RL agent need to be adjusted.
Restraining the actions of the RL agent as little as possible is desirable to avoid a negative impact on the learning performance.

If the obstacles in the environment do not move deterministically but according to a stochastic model, only stochastic statements about potential collision risks are possible. 
As a consequence, a trade-off between the exploration of the environment and the risk of a collision has to be made. 
In our work, we consider deterministic and stochastic environments with and without safe resting states.

\section{Approach}
\begin{figure*}[t]
    \vspace{-0.0cm}
    \input{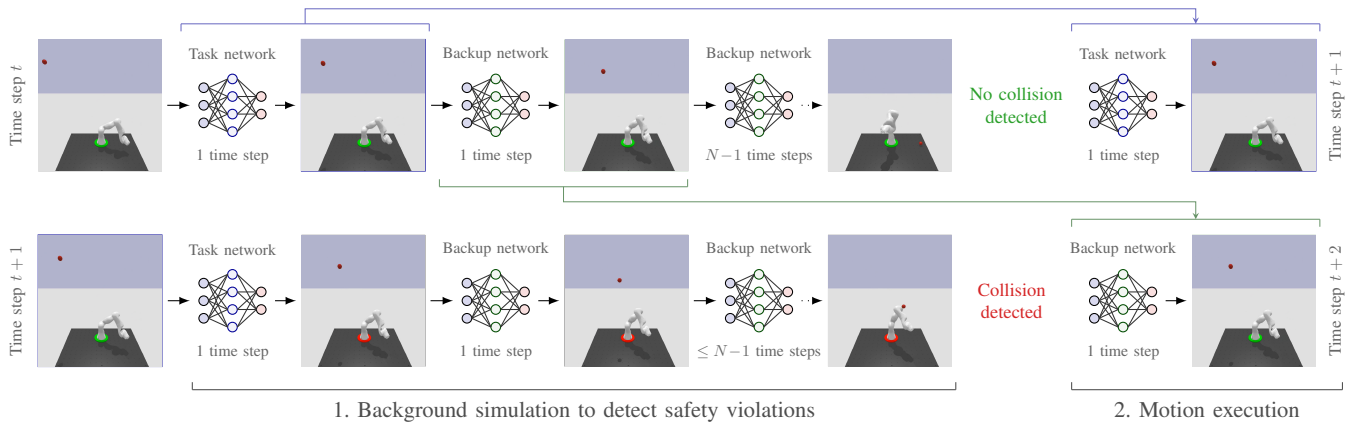}
    \vspace{-0.0cm}
	\caption{%
    Collision avoidance by ensuring the existence of a safe backup trajectory. See section \ref{sec:basic_principle} for details. 
    }
	\label{fig:basic_principle}
	\vspace{-0.4cm}
\end{figure*}
\subsection{Basic principle}
\label{sec:basic_principle}

The basic principle of our approach is illustrated in Fig.~\ref{fig:basic_principle}.
The figure shows two time steps of an environment in which a robot has to avoid a ball thrown in its direction. %
Our goal is to train a task policy, represented by a so-called \textit{task network}, without causing a collision. 
For this purpose, we make use of a backup policy, represented by a \textit{backup network},  that was trained to avoid collisions beforehand. 
At both time steps $t$ and $t+1$, the first step is to compute a desired action using the task network. Given the current state of the environment as input, the output of the task network is a probability distribution from which a desired action is sampled. 
However, before executing the motion resulting from the selected action, the backup network is used to generate a backup trajectory with a duration of $N$ time steps.
The composed movement with a duration of $N\!+\!1$ time steps is then checked for collisions in a background simulation using a physics simulator.
In the case of time step $t$, no collision is found during the background simulation. Consequently, the motion from time step $t$ to $t+1$ is executed as defined by the action from the task network. Contrary to this, a collision is detected during the background simulation conducted at time step $t+1$. Thus, the action from the task network is replaced by an action from the backup network.
Note that the resulting motion from time step $t+1$ to $t+2$ was part of the collision-free trajectory simulated at time step~$t$. 

\newpage
\noindent
Provided that
\begin{itemize}
    \item the background simulation accurately reflects reality and 
    \item a collision-free backup trajectory is found at least every $N+1$ time steps,
\end{itemize}
the method described above ensures that the resulting motion of the robot is always collision-free.
Ideally, $N$ would be infinite. 
In practice, $N$ is limited by the computational effort required for the background simulation. 
If obstacles do not move deterministically, a single background simulation is no longer sufficient to decide if an action is safe.
To account for stochastic environments, a so-called risk network can be trained to predict the probability of a collision based on data from previous rollouts of the backup policy. 
In this case, a trade-off between the amount of adjusted actions and the resulting average time to a collision emerges. The trade-off can be controlled by selecting a suitable threshold value.

In the remainder of this section, we analyze potential failure causes (\ref{sec:failure_causes}), describe how the backup network is trained (\ref{sec:training_backup_network}), clarify the risk estimation via risk networks (\ref{sec:risk_network}) and explain the training of the task network~(\ref{sec:risk_network}). 

\subsection{Failure mode analysis}
\label{sec:failure_causes}
\begin{figure}[t]
    \vspace{-0.3cm}
    \input{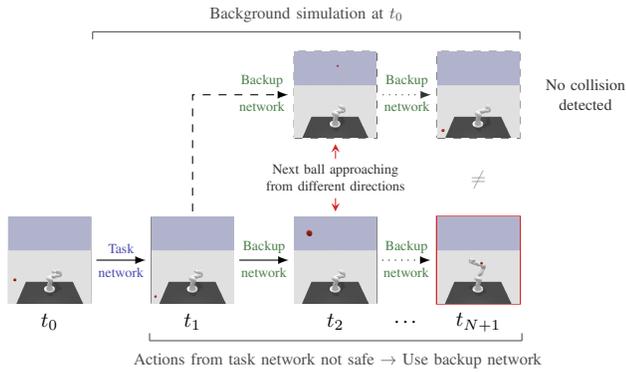}
    \vspace{-0.42cm}
	\caption{%
    Potential failure causes when performing a background simulation to detect safety violations.}
	\label{fig:failure_causes}
	\vspace{-0.5cm}
\end{figure}
In the following, we analyze conditions that can lead to a collision even if a background simulation is performed. 
As shown in Fig. \ref{fig:failure_causes}a, a collision can occur if the initial state of the environment is not safe. 
More precisely, this means that a rollout of the backup policy from the initial state leads to a collision.  
Collisions can also occur if the time horizon of the background simulation is too short (Fig.~\ref{fig:failure_causes}b).
A potential mitigation is to extend the time horizon of the background simulation~$N$.
However, a longer time horizon increases the computational effort required for the background simulation. 
A third possible failure cause becomes apparent when the environment is non-deterministic. 
In the example shown in Fig.~\ref{fig:failure_causes}c, a new ball is sampled once the previous ball missed the robot. %
The direction of the new ball is selected randomly and is therefore not known in advance. 
As a result, the ball direction selected during the background simulation differs from the actual ball direction. 
In stochastic environments, it is possible to estimate the probability of a collision based on data from previous rollouts of the backup policy. 
This can be done be training a so-called risk network via supervised learning. 
When using a risk network, the computational effort no longer depends on the time horizon~$N$, so that real-time execution becomes possible even if a long time horizon needs to be taken into account. 
Risk networks can also be used to estimate the risk of initial states. 
On the downside, it is important to note that incorrect risk predictions by the risk network introduce an additional source of error.

\vspace{-0.2cm}
\subsection{Training of the backup policy}
\label{sec:training_backup_network}
To train the backup policy~$\pi_B$ via model-free RL, we define a Markov decision process $(\mathcal{S}, \mathcal{A}, P, R)$.
The backup policy is represented by a neural network trained to map states $s_t \in \mathcal{S}$ to actions $a_t \in \mathcal{A}$ such that the sum of future rewards is maximized. 

\subsubsection{State space $\mathcal{S}$} \label{sec:backup_state}
Each state $s_t \in \mathcal{S}$ is composed of two parts: 
The first part $s_{t_{K\!i}}$ describes the kinematic state of the robot, and the second part $s_{t_{M\!o}}$ defines the state of the moving obstacles in the environment. More precisely, $s_{t_{K\!i}}$~consists of the current position $\mathrm{p}_t$, velocity $\mathrm{v}_t$ and acceleration $\mathrm{a}_t$ of each controllable robot joint, normalized to their respective maximum values. Details on the environment-specific part~$s_{t_{M\!o}}$ are given in section~\ref{sec:evaluation_environments}.
\subsubsection{Action space $\mathcal{A}$} 
\label{sec:action_space}
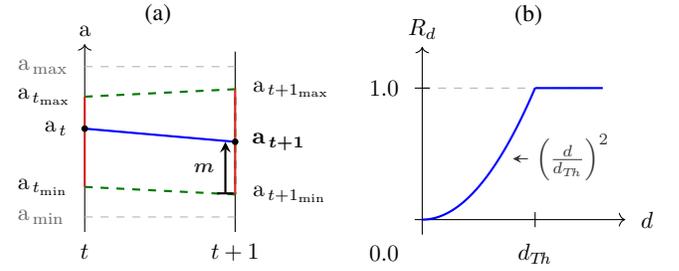
\begin{figure}[t]
\vspace{-0.2cm}
\captionsetup[subfigure]{margin=0pt}
    \vspace{0.21cm}
    \begin{subfigure}[c]{0.23\textwidth}
	   \vspace{-0.0cm}
           \hspace*{0.2cm}
	\def\ymax{1.3}
	\def\ymin{-1.2}
	\def\xdelta{2.0}
	\def\xmax{3*\xdelta + 0.5}
	\def\amin{-1.0}
	\def\amax{1.0}
	
	\def\aonemaxA{1.3}
	\def\aonemaxB{1.1}
	\def\jmin{-2.6} 
	\def\aonemaxC{\amin-\jmin}
	\def\astarB{-2.0}
	\def\astarC{-0}
	\def\jmax{0.65}
	\def\njmaxFractionC{0.75}
	\def\vstarminusoneA{1.2}
	\def\vstarminusoneB{1.0}
	\def\vstarminusoneC{0.8}
	\def\vstarA{-1.2}
	\def\vstarB{-0.4}
	\def\vstarC{0.0}
	\def\tfive{(\x - 5*\xdelta) / \xdelta}
	
	\def\jmaxVis{\jmax*3}
	\def\astarBVis{\astarB*1}
	\def\astarCVis{\astarC*1}
	
	\def\cfiveB{(\vstarminusoneB - \vstarB) / (0.5 * \jmaxVis - \astarBVis)}
	
	\def\dfiveB{\vstarB * (1-\cfiveB)}
	
	\def\cfiveC{(\vstarminusoneC - \vstarC) / (0.5 * \jmaxVis - \astarCVis)}
	\def\dfiveC{\vstarC * (1-\cfiveC)}
	
	\def\steps{1}
	\def\showLimits{1}
	\def\violation{0}
	\def\showRedLine{1}
	\def\showAcc{1}
	
	\def\redLineLinestyle{solid}
    \def\limitsLinestyle{dashed}
	
	\def\azeroMax{0.6}
	\def\aoneMax{0.7}
	\def\atwoMax{0.9}
	\def\athreeMax{1.2}
	\def\afourMax{1.6}
	
	\def\azeroMin{-0.6}
	\def\aoneMin{-0.7}
	\def\atwoMin{-1.3}
	\def\athreeMin{-1.5}
	\def\afourMin{\amin}
	
	\def\azero{0.175}
	
	\def\actionColor{black!85}
	
	\def\aoneViolation{1.5}
	\ifnum \violation = 1
		\def\aone{\aoneViolation}
	\else
		\def\aone{0.5 * \aoneMin + 0.5 * \aoneMax}
	\fi
	
	\def\atwo{0.75 * \atwoMin + 0.25 * \atwoMax}
	\def\athree{1.0 * \athreeMax + 0.0 * \athreeMin}
	\def\afour{1.2}

	\definecolor{POS_LIM_A}{RGB}{0,120,0}%
	\definecolor{POS_LIM_B}{RGB}{0,0,150} %
	\definecolor{POS_LIM_C}{RGB}{0,120,0}%
	\definecolor{POS_LIM_D}{RGB}{214,39,40}
     \def\legendScale{0.88}
	
	\begin{tikzpicture}[scale=1, every node/.style={scale=\legendScale}]
	\draw [<-] (0,\ymax) node (yaxis) [above] {$\mathrm{a}$} -- (0,\ymin) node[yshift=-0.3cm, name=nodet0] {$t$};
	
	\draw[] (\xdelta, \ymax - 0.1)  -- (\xdelta, \ymin) node[yshift=-0.3cm] {$t+1$};

	\ifnum \showLimits > 0
		\draw[\limitsLinestyle, color=POS_LIM_A, thick] (0,\azeroMax) node[left=0.07cm, color=black] {$\mathrm{a}_{\,t_{\mathrm{max}}}$} -- (\xdelta,\aoneMax)  node[pos=1.0, right=0.12cm, color=black] {$\mathrm{a}_{\,t+1_{\mathrm{max}}}$} -- (\xdelta,\aoneMin); 
		\ifnum \steps > 1
			\draw[\limitsLinestyle, color=POS_LIM_A, very thick] (\xdelta,\aoneMax) -- (2*\xdelta,\atwoMax); 
			\ifnum \steps > 2
				\draw[\limitsLinestyle, color=POS_LIM_A, very thick] (2*\xdelta,\atwoMax) -- (3*\xdelta,\athreeMax); 
				\ifnum \steps > 3
					\draw[\limitsLinestyle, color=POS_LIM_A, very thick] (3*\xdelta,\athreeMax) -- (4*\xdelta,\afourMax);
					\fi
			\fi
		\fi
		\draw[\limitsLinestyle, color=POS_LIM_C, thick] (0,\azeroMin) node[left=0.105cm, color=black] {$\mathrm{a}_{\,t_{\mathrm{min}}}$} -- (\xdelta,\aoneMin) node[pos=1.0, right=0.12cm, color=black] {$\mathrm{a}_{\,t+1_{\mathrm{min}}}$} -- (\xdelta,\aoneMin); 
		\ifnum \steps > 1
			\draw[\limitsLinestyle, color=POS_LIM_C, very thick] (\xdelta,\aoneMin) -- (2*\xdelta,\atwoMin); 
			\ifnum \steps > 2
				\draw[\limitsLinestyle, color=POS_LIM_C, very thick] (2*\xdelta,\atwoMin) -- (3*\xdelta,\athreeMin); 
				\ifnum \steps > 3
					\draw[\limitsLinestyle, color=POS_LIM_C, very thick] (3*\xdelta,\athreeMin) -- (4*\xdelta,\afourMin);
				\fi
			\fi
		\fi
	\fi
	\ifnum \showRedLine = 1
	    \draw[\redLineLinestyle, color=POS_LIM_D,  thick] (0,\azeroMin)  -- (0,\azeroMax);
		\draw[\redLineLinestyle, color=POS_LIM_D,  thick] (\xdelta,\aoneMin)  -- (\xdelta,\aoneMax);
	\fi

	\ifnum \showAcc = 1
		\ifnum \violation = 0
            \draw[color=blue, thick] (0,\azero)  -- (\xdelta,\aone) node[pos=1.0, right=0.19cm, color=black, name=a_t_plus_one_n, outer sep=0pt, inner sep=1.25pt]{$\boldsymbol{\mathrm{a}_{\,t+1}}$};

			\fill[black] (\xdelta,\aone) circle (1.25pt);
    		\draw[|-stealth, thick, draw=black]  ($(\xdelta, \aoneMin) + (-0.065*\xdelta, 0)$) -- ($(\xdelta, \aone) + (-0.065*\xdelta, 0)$) node[pos=0.5, left=0.05cm, color=\actionColor, scale=0.85] {$\boldsymbol{m}$};
		\else
			\draw[color=POS_LIM_B, very thick, dashed] (0,\azero)  -- (\xdelta,\aoneViolation) node[pos=0, left=0.15cm, color=black] {$a_0$}; %
		\fi
		\ifnum \steps > 1
			\ifnum \violation = 0
				\draw[color=POS_LIM_B, very thick] (\xdelta,\aone) -- (2*\xdelta,\atwo) node[pos=1.0, above=-0.53cm, color=black, xshift=0.65cm, name=a_t_plus_two, outer sep=0pt, inner sep=1.75pt]{};%
    			\fill[black] (2*\xdelta,\atwo) circle (1.5pt);
        		\draw[|-stealth, thick, draw=\actionColor]  ($(2*\xdelta, \atwoMin) + (-0.065*\xdelta, 0)$) -- ($(2*\xdelta, \atwo) + (-0.065*\xdelta, 0)$) node[pos=0.65, left=0.05cm, color=\actionColor, scale=0.9] {$\underline{a}_{t+1}$};
			\else
				\draw[color=POS_LIM_B, very thick, dashed] (\xdelta,\aoneViolation) -- (2*\xdelta,\atwo); 
			\fi
			
			\ifnum \steps > 2
				\draw[color=POS_LIM_B, very thick] (2*\xdelta,\atwo) -- (3*\xdelta,\athree); 
				\fill[black] (3*\xdelta,\athree) circle (1.5pt);
        		\draw[|-stealth, thick, draw=\actionColor]  ($(3*\xdelta, \athreeMin) + (-0.065*\xdelta, 0)$) -- ($(3*\xdelta, \athree) + (-0.065*\xdelta, -0.2)$) node[pos=0.5, left=0.05cm, color=\actionColor, scale=0.9] {$\underline{a}_{t+2}$};
				\ifnum \steps > 3
					\draw[color=POS_LIM_B, very thick] (3*\xdelta,\athree) -- (4*\xdelta,\afour);
				\fi
			\fi
		\fi
	\fi

	\fill[black] (0,\azero) circle (1.25pt) node[above=\azero, left=0.12cm, color=black] {$\mathrm{a}_{\,t}$};

	\draw[dashed, draw=black!30] (0, \amin) node[left=0.16cm, black!60] {$\mathrm{a}_{\,\mathrm{min}}$}  -- (\xdelta, \amin) ;
	\draw[dashed, draw=black!30] (0, \amax) node[left=0.11cm, black!60] {$\mathrm{a}_{\,\mathrm{max}}$}  -- (\xdelta, \amax) ;

	\end{tikzpicture} 

	   \vspace{-4.6cm}\hspace*{-0.53cm}\subcaptionbox{}[5cm]

	\end{subfigure}
	\begin{subfigure}[c]{0.23\textwidth}
	    \vspace{-0.08cm}
	    \hspace*{0.3cm}
\vspace*{0.4cm}
\begin{tikzpicture}
  \def\legendScale{0.88}
  \def\tickLength{0.2}
  \def\lStateX{1.5}
  \def\lEndX{2.4}
  \def\yOne{1.75}
  \def\yMax{2.3}
  \def\annotationOffset{0.4}
  \draw[dashed, draw=black!30] (0, \yOne) -- (\lEndX, \yOne);
  \draw[->] (-0.1, 0) -- (2.7, 0) node[pos=0, left=0.1cm, yshift=-0.45cm, scale=\legendScale]{$0.0$} node[right=0.1cm, scale=\legendScale] {$d$};
  \draw[->] (0, -0.2) -- (0, \yMax) node[above, scale=\legendScale] {$R_d$};
  \draw (-0.5*\tickLength, \yOne) -- + (\tickLength, 0) node[pos=0, left=0.1cm, scale=\legendScale]{$1.0$};
  \draw[scale=1.0, domain=-0:\lStateX, smooth, variable=\x, blue, thick] plot ({\x}, {\x*\x*\yOne*(1/(\lStateX*\lStateX))});
  \draw[scale=1.0, domain=\lStateX:\lEndX, smooth, variable=\x, blue, thick] plot ({\x},
  {(\yOne)});
  \draw (\lStateX, -0.5*\tickLength) -- + (0, \tickLength)  node[pos=0, yshift=-0.35cm, scale=\legendScale](d_threshold){$d_{T\!h}$};
  \node[above of=d_threshold, node distance=1.28cm, xshift=0.5cm, color=black!80, scale=\legendScale](equation_description){$\left(\frac{d}{d_{T\!h}}\right)^2$};
  \draw[-stealth, color=black!80, thin]($(equation_description.182) + (-0.0cm, 0)$) -- + (-0.2cm, 0);

\end{tikzpicture}

		\vspace{-5.03cm}\hspace*{0.1cm}\subcaptionbox{}[5cm]

	\end{subfigure} 
	\vspace{3.2cm}
	\caption{Action mapping (a) and distance reward (b).}
	\label{fig:action_mapping_and_reward}
	\vspace{-0.4cm}
\end{figure}

We utilize an action space introduced in~\cite{kiemel2021learning}, which ensures that the kinematic joint limits \mbox{(\ref{eq:constraint_p}) - (\ref{eq:constraint_j})} are satisfied at all times. 
An action $a_t \in \mathcal{A}$ consists of one scalar $m \in [-1, 1]$ per controllable robot joint. As shown in Figure~\ref{fig:action_mapping_and_reward}a, the scalar~$m$ specifies a joint acceleration $\mathrm{a}_{t+1}$  within a range of feasible accelerations $[\mathrm{a}_{t+1_{\mathrm{min}}}, \mathrm{a}_{t+1_{\mathrm{max}}}]$.
To generate continuous acceleration setpoints, the current acceleration $\mathrm{a}_{t}$ is linearly connected to $\mathrm{a}_{t+1}$. 
Subsequently, velocity and position setpoints for a trajectory controller are computed through integration. 
By enforcing jerk limits for each robot joint, we ensure that the resulting trajectory is smooth no matter which action is selected. 

\subsubsection{Reward function $R$}
We compute the immediate \mbox{reward $r_t$} for an action $a_t$ as follows:
\begin{align}
r_t = \alpha \cdot R_{d_{{t+1}_{M\!o}}} + \beta  \cdot R_{d_{{t+1}_{S\!t}}} + \gamma  \cdot R_{d_{{t+1}_{S\!c}}} + R_{T_B}, 
\end{align}\vspace*{-0.375cm}

\noindent
where $\alpha$, $\beta$ and $\gamma$ are weighing factors.
The reward components  $R_{d_{{t+1}_{M\!o}}}$, $R_{d_{{t+1}_{S\!t}}}$ and $R_{d_{{t+1}_{S\!c}}}$ depend on $d_{{t+1}_{M\!o}}$, $d_{{t+1}_{S\!t}}$ and $d_{{t+1}_{S\!c}}$, the minimum distances to moving obstacles, static obstacles and self-collisions at $t+1$.
The shape of the function used to compute $R_{d_{{t+1}_{M\!o}}}$, $R_{d_{{t+1}_{S\!t}}}$ and $R_{d_{{t+1}_{S\!c}}}$ is shown in Fig.~\ref{fig:action_mapping_and_reward}b.
In the figure, the variable~$d$ corresponds to $d_{{t+1}_{M\!o}}$, $d_{{t+1}_{S\!t}}$ or $d_{{t+1}_{S\!c}}$, while $d_{T\!h}$ is a fixed threshold value. 
For distances smaller than $d_{T\!h}$, the reward increases as~$d$ increases.
As a result, $R_{d_{{t+1}_{M\!o}}}$, $R_{d_{{t+1}_{S\!t}}}$ and $R_{d_{{t+1}_{S\!c}}}$ 
encourage the robot to keep a distance from moving obstacles, static obstacles and self-collisions, respectively.
The last term $R_{T_B}$ is a termination bonus which is always zero except when the episode terminates without a collision.

\subsubsection{Termination}
An episode is terminated after $T$ time steps or earlier if a collision occurs.
Note that the immediate reward is never negative, discouraging an early termination.  
\subsubsection{Sampling of initial states}
\label{sec:initial_states_sampling}
Once trained, we do not update the backup policy. For that reason, it is important to cover a wide range of initial states during the training of the policy.  
To do so, the moving obstacles are initialized in a random configuration. 
After that, random joint positions are sampled until a collision-free robot position is found. To find feasible initial joint velocities and accelerations, we first choose random values within the specified kinematic limits (\ref{eq:constraint_v}) - (\ref{eq:constraint_a}) and compute the range of feasible joint accelerations $[\mathrm{a}_{\,t+1_{\mathrm{min}}}, \mathrm{a}_{\,t+1_{\mathrm{max}}}]$ as explained in section \mbox{\ref{sec:action_space}}.
If the range is empty, the sampling process is repeated. %

\def\riskSignal{\ensuremath{c_t}}
\def\riskThreshold{c_{T\!h}}

\subsection{Data-based risk estimation}
\label{sec:risk_network}
\subsubsection{Data generation}
\label{sec:risk_network_data_generation}
The first step towards training a data-based risk estimator is to generate a dataset from rollouts of the backup policy. 
For that purpose, we utilize a physics simulator and initialize the environment in a random state~$s_t$ as described in section \ref{sec:initial_states_sampling}.
We then select a uniformly random action $a_t$ and simulate the next time step until $s_{t+1}$.
Starting from $s_{t+1}$, a rollout of the backup policy with $N$ time steps is simulated. If a collision occurs during the $N+1$ simulated time steps, a risk signal $\riskSignal$ is defined to be $1.0$, otherwise $0.0$.
For each rollout, we store the \mbox{tuple ($s_t$, $a_t$, $s_{t+1}$, $\riskSignal$)}.
\vspace{0.1cm}
\subsubsection{Training of risk networks}
We evaluate two different types of risk networks, both trained with a binary cross-entropy loss using supervised learning. 
State-action-based risk networks are trained to predict the risk $\riskSignal$ given $s_t$ and $a_t$, whereas state-based risk networks use $s_{t+1}$ to predict~$\riskSignal$.
The use of data-based risk estimators makes it possible to account for the stochastic behavior of moving obstacles. 
Specifically, a risk network can learn to output a risk prediction between $0.0$ and $1.0$ depending on the probability of a collision.

\subsection{Risk-aware training of the task policy} %
\label{sec:training_task_network}
As for the backup policy, the training of the task policy is based on model-free RL and a  Markov decision process. A state $s_t \in \mathcal{S}$ consists of three parts  $s_{t_{K\!i}}$, $s_{t_{M\!o}}$ and $s_{t_{T\!a}}$. 
While $s_{t_{K\!i}}$ and $s_{t_{M\!o}}$ correspond to the state components used for the backup network, the third part $s_{t_{T\!a}}$ encodes additional task-specific information.
An action $a_t \in \mathcal{A}$ is composed of $a_{t_{K\!i}}$ and $a_{t_{T\!a}}$, where $a_{t_{K\!i}}$ determines the movement of the robot joints and $a_{t_{T\!a}}$ is an additional task-specific action component. %
If the risk~$\riskSignal$ predicted by the risk network exceeds a predefined risk threshold $\riskThreshold$, the action component $a_{t_{K\!i}}$ is replaced by an action from the backup network, which is denoted as~$a^{B}_{t_{K\!i}}$. 
Since both the task network and the backup network utilize the same method to map actions to movements, the kinematic joint limits \mbox{(\ref{eq:constraint_p}) - (\ref{eq:constraint_j})} are satisfied no matter which policy is executed. The reward function for the training of the task policy depends on the desired learning task. A concrete example for a reaching task is given by equation~(\ref{eq:reward_reaching_task}) in section~\ref{sec:evaluation_risk_networks}.

\begin{figure}[t]
    \vspace{0.1cm}
    \input{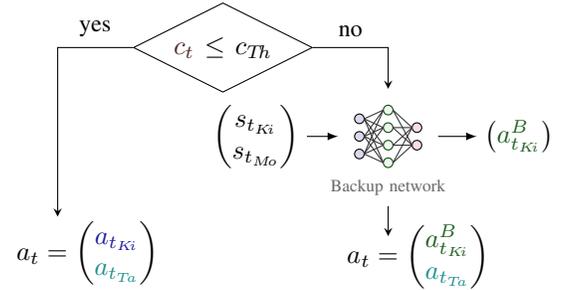}
    \vspace{-0.25cm}
	\caption{
    The figure illustrates the action generation with the task network, four different ways (A, B1, B2a, and B2b) to estimate the corresponding risk, and the risk-dependent action adjustment using the backup network.}
	\label{fig:action_adaptation}
	\vspace{-0.9cm}
\end{figure}

The procedure of the risk-dependent action adjustment is shown in Fig. \ref{fig:action_adaptation}. We analyze and compare four different methods to estimate the risk~$\riskSignal$. The first option~(A) is to use the state-action-based risk network. %
The other three options rely on the state-based risk network. 
In the case of (B1), the current state is used for the risk prediction. 
A disadvantage of this method is that it tends to adjust actions too late, i.e. when the robot is already in a risky state. 
This drawback can be avoided by checking the risk of the following state $s_{t+1}$ instead. However, the state $s_{t+1}$ is usually not fully known in advance.
While the kinematic part $s_{{t+1}_{K\!i}}$ can be computed using the action from the task network, the state of the moving obstacles $s_{{t+1}_{M\!o}}$ is either considered constant (B2a) or forecasted (B2b), e.g. with a Kalman filter.
Note that when checking the risk of the following state $s_{t+1}$, the risk of the transition from $s_{t}$ to $s_{t+1}$ is not explicitly considered. %

\begin{figure}[!bp]
\captionsetup[subfigure]{margin=80pt}
    \vspace{0.1cm}
	
	    \begin{subfigure}[c]{0.23\textwidth}
	   \vspace{0.0cm} 
	   \includegraphics[trim=625 250 625 300, clip, width=\textwidth]{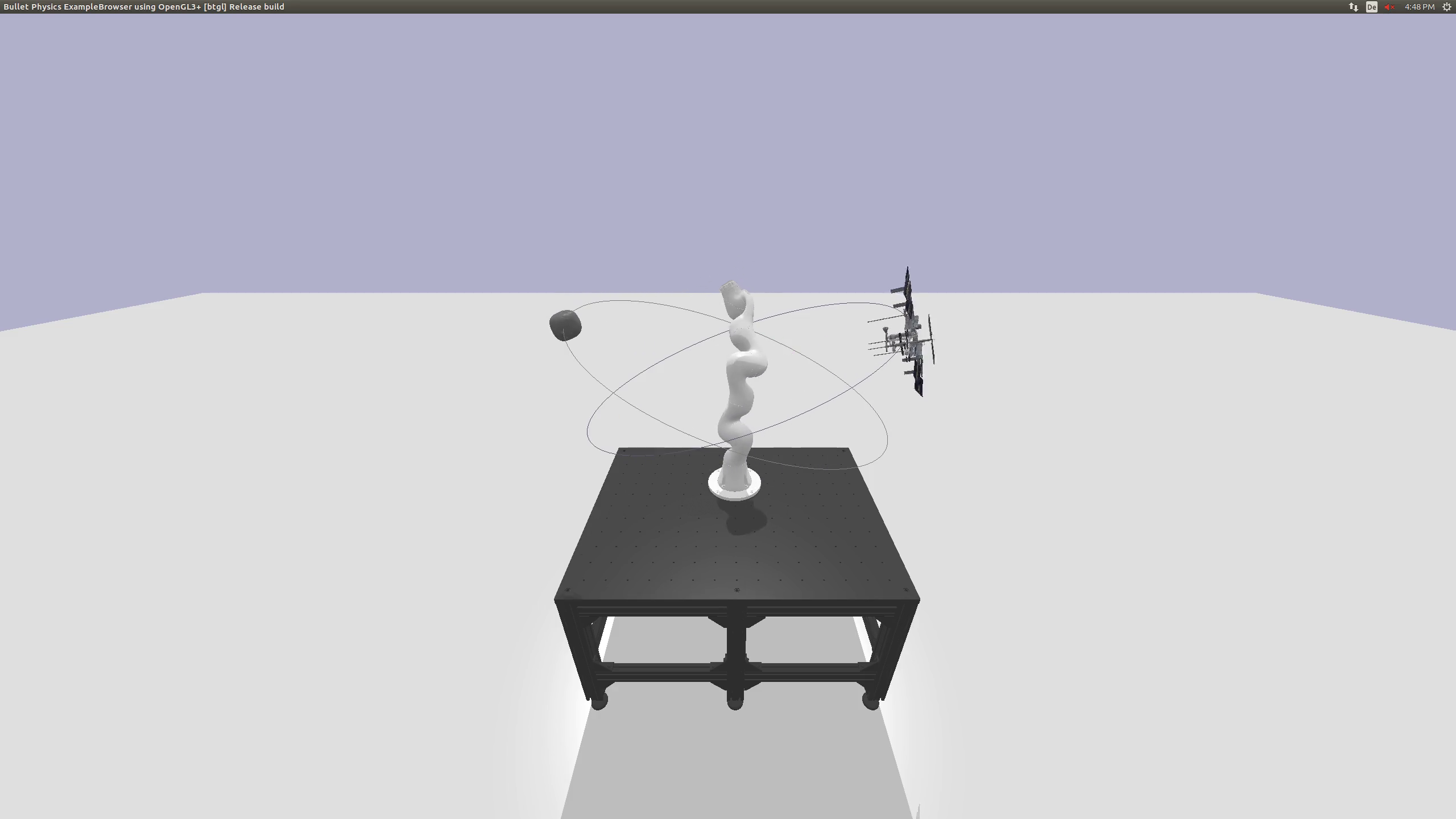}
	   
	   \vspace{-0.425cm}\hspace*{-2.05cm}\subcaptionbox{Space}[8cm]

	\end{subfigure}
	\hspace{0.0065\textwidth}
	\begin{subfigure}[c]{0.23\textwidth}
	    \vspace{0.0cm}
	    \includegraphics[trim=580 325 670 225, clip, width=\textwidth]{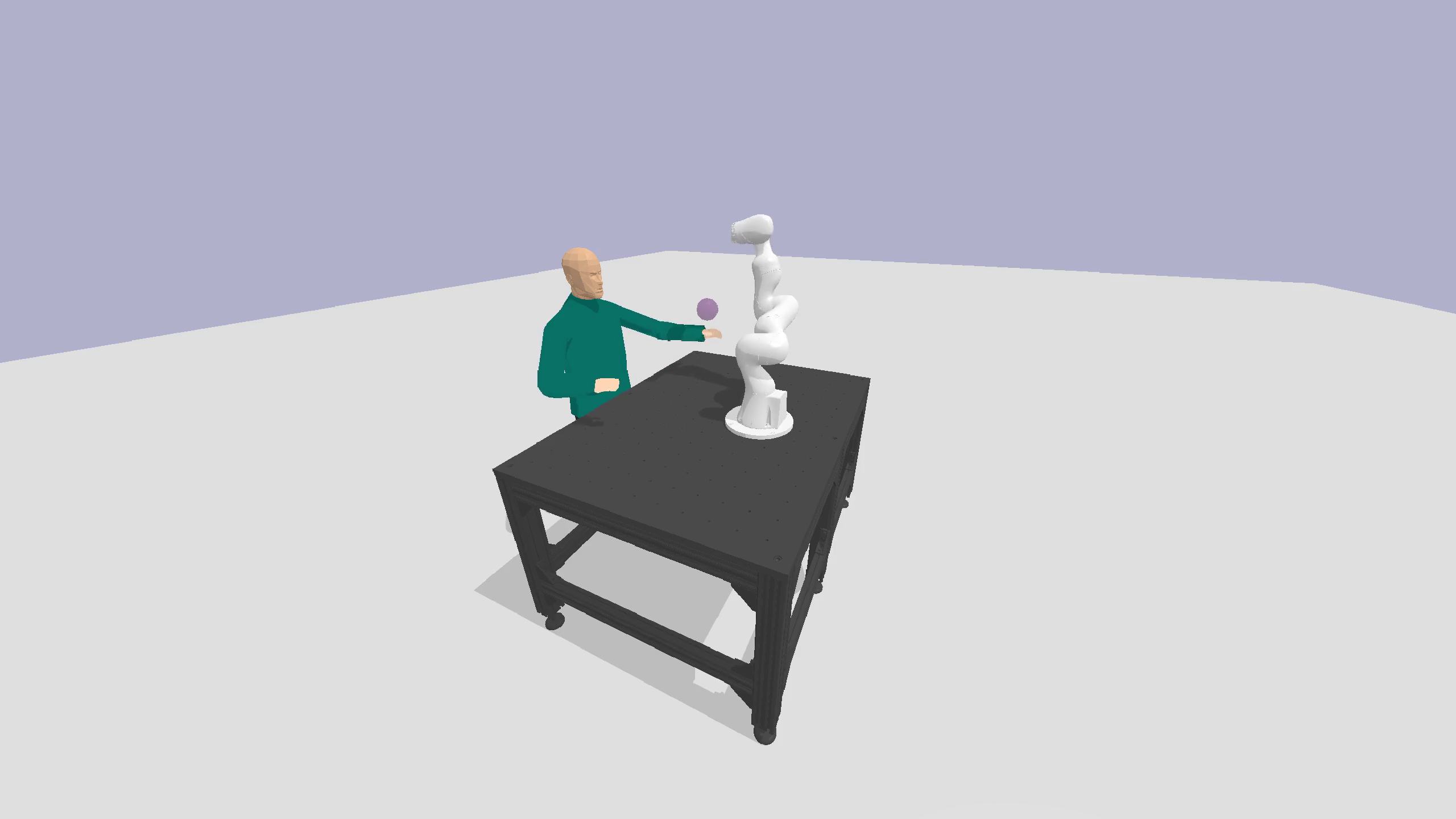}
	    
		\vspace{-0.425cm}\hspace*{-2.55cm}\subcaptionbox{Human}[9cm]

	\end{subfigure}

	\caption{%
Two environments used for our evaluation.}%
    \vspace{0.25cm}
	\label{fig:evaluation_environments}
\end{figure}
\newpage
\section{Evaluation}
\subsection{Evaluation environments}
\label{sec:evaluation_environments}
We evaluate our approach in three different environments: 
A space environment and a human environment, shown in Fig.~\ref{fig:evaluation_environments}, and a ball environment shown in Fig.~\ref{fig:three_steps}.
\subsubsection{Space (deterministic)}
A space station and an asteroid orbit the robot in a deterministic manner.
The environment-specific part of the state $s_{t_{M\!o}}$ indicates the position of both objects on their orbit. 
\subsubsection{Ball (stochastic)}
In this environment, a ball is thrown towards the robot. The environment is stochastic as the initial position and the direction of the ball are selected at random. 
Once the ball is thrown, it moves along a parabolic trajectory due to gravity.
The environment-specific part of the state $s_{t_{M\!o}}$ includes the position and the velocity of the ball.
A new ball is thrown, as soon as the previous one missed the robot. 
\subsubsection{Human (stochastic)}
In this environment, a human moves his hand to a target point without considering the motions of the robot. The environment is stochastic since the position of the target point is selected at random.
The state component $s_{t_{M\!o}}$ indicates the kinematic state of the human arms but not the desired \mbox{target point} of the human.

In all environments, a collision occurs on average after three seconds if a task policy selects random actions without making use of a backup policy (see Table~\ref{table:environments_metrics_random_agent}).

\begin{table}[h]
     \vspace{-0.1cm}
     \caption{Results for random actions without our method.}
     \vspace{-0.15cm}
    \makegapedcells
\begin{tabular*}{0.49\textwidth}{@{}p{16.0mm}p{12mm}p{10.0mm}p{12.0mm}p{20mm}} 
    \toprule
\hspace{0.02cm} Environment & \hfil Time until \hfil  & \hfil Self- \hfil &  \hfil Collision \hfil & \hfil Collision with \hfil \\
& \hfil collision \hfil & \hfil collision \hfil &  \hfil with table \hfil & \hfil moving obstacles \hfil \\
    \hline
\hspace{0.0002cm} \tabitem Space  & \hfil \hspace{0.000cm} \SI{2.6}{\s} &  \hspace{0.000cm} \hfil \hspace{0.000cm} \SI{30}{\percent} & \hfil \SI{30}{\percent} & \hfil \SI{40}{\percent} \\
\hspace{0.0002cm} \tabitem Ball & \hfil \hspace{0.000cm}  \SI{3.3}{\s}  & \hspace{0.000cm} \hfil  \hspace{0.000cm} \SI{38}{\percent} & \hfil \SI{45}{\percent} & \hfil \SI{17}{\percent} \\
\hspace{0.0002cm} \tabitem Human & \hfil \hspace{0.000cm} \SI{3.4}{\s} & \hspace{0.000cm} \hfil  \hspace{0.000cm} \SI{38}{\percent} & \hfil  \SI{41}{\percent} & \hfil \SI{21}{\percent} \\
    \bottomrule
    \end{tabular*}
    \vspace{0ex}
\label{table:environments_metrics_random_agent}
\vspace{-0.58cm}
\end{table}

\subsection{Training of the backup policy}
\label{sec:evaluation_backup_policy}
To train a backup policy for each environment, we use the on-policy RL-algorithm PPO \cite{schulman2017proximal} and a feedforward neural network with two hidden layers.
The time between decision steps is set to \SI{0.1}{\s}. During training, each episode is terminated after $T\!=\!20$ time steps or earlier if a collision occurs. As shown in Table~\ref{table:evaluation_backup_policy}, the fraction of collision-free episodes increases significantly during training. 
Note that the fraction cannot reach \SI{100}{\percent} since the robot is sometimes initialized in states where no evasive motion exists. 
The rightmost column of Table~\ref{table:evaluation_backup_policy} shows the time until a collision if the robot is initialized such that no collision occurs during the first \SI{2.0}{\s}. 
In the space and the human environment, very high values are achieved. 
In these environments, the backup policy can guide the robot into a region of the workspace where no \mbox{collisions with} moving obstacles occur.
In the ball environment \mbox{there is} no such region. As a result, a collision occurs on average after~\SI{85}{\s}.
\begin{table}[h]
    \vspace{-0.2cm}
     \caption{Training results of the backup policy.}
     \vspace{-0.15cm}
    \makegapedcells
\begin{tabular*}{0.49\textwidth}{@{}p{16.0mm}p{21mm}p{21.0mm}p{15.0mm}} 
    \toprule
\multirow[t]{4}{*}{\hspace{0.02cm} Environment}
 & \multicolumn{2}{c}{Episodes without collision within $\SI{2.0}{\s}$} & \multicolumn{1}{c}{Time until}  \\
& \multicolumn{1}{c}{Untrained agent}  & \multicolumn{1}{c|}{Trained agent} & \multicolumn{1}{c}{collision}  \\
    \hline
\hspace{0.0002cm} \tabitem Space  & \hfil \hspace{0.000cm} \SI{38}{\percent} &  \hspace{0.000cm} \hfil \hspace{0.000cm} \SI{91}{\percent} & \hfil $>\SI{40000}{\s}$ \\
\hspace{0.0002cm} \tabitem Ball & \hfil \hspace{0.000cm}  \SI{48}{\percent}  & \hspace{0.000cm} \hfil  \hspace{0.000cm} \SI{91}{\percent} & \hfil$\:\:\:\:\:\:$\SI{85.4}{\s} \\
\hspace{0.0002cm} \tabitem Human & \hfil \hspace{0.000cm} \SI{51}{\percent} & \hspace{0.000cm} \hfil  \hspace{0.000cm} \SI{94}{\percent} & \hfil $>\SI{40000}{\s}$ \\
    \bottomrule
    \end{tabular*}
    \vspace{0ex}
\label{table:evaluation_backup_policy}
\vspace{-0.6cm}
\end{table}

\begin{table}[t]
     \caption{Random actions with background simulations. \textcolor{TABLE_ENTRY_ONE}{Time until collision}, \textcolor{TABLE_ENTRY_TWO}{Action adjustment rate}, \textcolor{TABLE_ENTRY_THREE}{Collision with table or self-collision}, \textcolor{TABLE_ENTRY_FOUR}{Computation time per simulation time}}
      \vspace{-0.15cm}
    \makegapedcells
\begin{tabular*}{0.49\textwidth}{@{}p{20.0mm}p{8.5mm}p{8.5mm}p{8.5mm}p{10mm}p{10mm}} 
    \toprule
\hspace{0.02cm} Environment & \hfil $N=0$ \hfil  & \hfil $N=1$ \hfil &  \hfil $N=5$ \hfil & \hfil $N=20$ \hfil & \hfil $N=30$ \hfil \\
& \hfil \SI{0.0}{\s} \hfil & \hfil \SI{0.1}{\s} \hfil &  \hfil \SI{0.5}{\s} \hfil & \hfil \SI{2.0}{\s} \hfil & \hfil \SI{3.0}{\s} \hfil \\
    \hline
\hspace{0.02cm}    Space & \hfil  \textcolor{TABLE_ENTRY_ONE}{\SI{2.7}{\s}} &  \hfil  \textcolor{TABLE_ENTRY_ONE}{\SI{6.8}{\s}} & \hfil \textcolor{TABLE_ENTRY_ONE}{\SI{170.4}{\s}} & \hfil \textcolor{TABLE_ENTRY_ONE}{$>$\SI{2000}{\s}} & \hfil \textcolor{TABLE_ENTRY_ONE}{$>$\SI{2000}{\s}} \\
 & \hfil  \textcolor{TABLE_ENTRY_TWO}{\SI{4.3}{\percent}} &  \hfil \textcolor{TABLE_ENTRY_TWO}{\SI{6.7}{\percent}} & \hfil \textcolor{TABLE_ENTRY_TWO}{\SI{7.5}{\percent}} & \hfil \textcolor{TABLE_ENTRY_TWO}{\SI{7.5}{\percent}} & \hfil \textcolor{TABLE_ENTRY_TWO}{\SI{7.5}{\percent}} \\
 & \hfil  \textcolor{TABLE_ENTRY_THREE}{\SI{57}{\percent}} &  \hfil \textcolor{TABLE_ENTRY_THREE}{\SI{32}{\percent}} & \hfil \textcolor{TABLE_ENTRY_THREE}{\SI{0}{\percent}} & \hfil \textcolor{TABLE_ENTRY_THREE}{\SI{0}{\percent}} & \hfil \textcolor{TABLE_ENTRY_THREE}{\SI{0}{\percent}} \\
  & \hfil  \textcolor{TABLE_ENTRY_FOUR}{\SI{38}{\percent}} &  \hfil \textcolor{TABLE_ENTRY_FOUR}{\SI{48}{\percent}} & \hfil \textcolor{TABLE_ENTRY_FOUR}{\SI{119}{\percent}} & \hfil \textcolor{TABLE_ENTRY_FOUR}{\SI{235}{\percent}} & \hfil \textcolor{TABLE_ENTRY_FOUR}{\SI{303}{\percent}} \\
\hspace{0.02cm}    Ball & & & & &\\
\hspace{0.0002cm} \tabitem Stochastic$\!\!\!\!\!\!\!$  & \hfil  \textcolor{TABLE_ENTRY_ONE}{\SI{3.6}{\s}} &  \hfil  \textcolor{TABLE_ENTRY_ONE}{\SI{10.4}{\s}} & \hfil \textcolor{TABLE_ENTRY_ONE}{\SI{64.1}{\s}} & \hfil \textcolor{TABLE_ENTRY_ONE}{\SI{160.3}{\s}} & \hfil \textcolor{TABLE_ENTRY_ONE}{\SI{116.3}{\s}} \\
 & \hfil  \textcolor{TABLE_ENTRY_TWO}{\SI{3.4}{\percent}} &  \hfil \textcolor{TABLE_ENTRY_TWO}{\SI{4.7}{\percent}} & \hfil \textcolor{TABLE_ENTRY_TWO}{\SI{5.5}{\percent}} & \hfil \textcolor{TABLE_ENTRY_TWO}{\SI{7.1}{\percent}} & \hfil \textcolor{TABLE_ENTRY_TWO}{\SI{8.4}{\percent}} \\
  & \hfil  \textcolor{TABLE_ENTRY_THREE}{\SI{78}{\percent}} &  \hfil \textcolor{TABLE_ENTRY_THREE}{\SI{43}{\percent}} & \hfil \textcolor{TABLE_ENTRY_THREE}{\SI{6}{\percent}} & \hfil \textcolor{TABLE_ENTRY_THREE}{\SI{4}{\percent}} & \hfil \textcolor{TABLE_ENTRY_THREE}{\SI{9}{\percent}} \\
    & \hfil  \textcolor{TABLE_ENTRY_FOUR}{\SI{27}{\percent}} &  \hfil \textcolor{TABLE_ENTRY_FOUR}{\SI{42}{\percent}} & \hfil \textcolor{TABLE_ENTRY_FOUR}{\SI{119}{\percent}} & \hfil \textcolor{TABLE_ENTRY_FOUR}{\SI{239}{\percent}} & \hfil \textcolor{TABLE_ENTRY_FOUR}{\SI{290}{\percent}} \\
\hspace{0.0002cm} \tabitem Deterministic$\!\!\!\!\!\!\!$ & \hfil  \textcolor{TABLE_ENTRY_ONE}{\SI{3.8}{\s}} &  \hfil  \textcolor{TABLE_ENTRY_ONE}{\SI{10.9}{\s}} & \hfil \textcolor{TABLE_ENTRY_ONE}{\SI{68.7}{\s}} & \hfil \textcolor{TABLE_ENTRY_ONE}{\SI{164.7}{\s}} & \hfil \textcolor{TABLE_ENTRY_ONE}{\SI{181.6}{\s}} \\
 & \hfil  \textcolor{TABLE_ENTRY_TWO}{\SI{3.2}{\percent}} &  \hfil \textcolor{TABLE_ENTRY_TWO}{\SI{4.7}{\percent}} & \hfil \textcolor{TABLE_ENTRY_TWO}{\SI{5.5}{\percent}} & \hfil \textcolor{TABLE_ENTRY_TWO}{\SI{7.5}{\percent}} & \hfil \textcolor{TABLE_ENTRY_TWO}{\SI{8.5}{\percent}} \\
 & \hfil  \textcolor{TABLE_ENTRY_THREE}{\SI{82}{\percent}} &  \hfil \textcolor{TABLE_ENTRY_THREE}{\SI{43}{\percent}} & \hfil \textcolor{TABLE_ENTRY_THREE}{\SI{0}{\percent}} & \hfil \textcolor{TABLE_ENTRY_THREE}{\SI{0}{\percent}} & \hfil \textcolor{TABLE_ENTRY_THREE}{\SI{0}{\percent}} \\
   & \hfil  \textcolor{TABLE_ENTRY_FOUR}{\SI{27}{\percent}} &  \hfil \textcolor{TABLE_ENTRY_FOUR}{\SI{43}{\percent}} & \hfil \textcolor{TABLE_ENTRY_FOUR}{\SI{103}{\percent}} & \hfil \textcolor{TABLE_ENTRY_FOUR}{\SI{235}{\percent}} & \hfil \textcolor{TABLE_ENTRY_FOUR}{\SI{316}{\percent}} \\
\hspace{0.02cm}    Human & & & & &\\
\hspace{0.0002cm} \tabitem Stochastic$\!\!\!\!\!\!\!$  & \hfil  \textcolor{TABLE_ENTRY_ONE}{\SI{3.7}{\s}} &  \hfil  \textcolor{TABLE_ENTRY_ONE}{\SI{11.1}{\s}} & \hfil \textcolor{TABLE_ENTRY_ONE}{\SI{62.9}{\s}} & \hfil \textcolor{TABLE_ENTRY_ONE}{\SI{72.3}{\s}} & \hfil \textcolor{TABLE_ENTRY_ONE}{\SI{64.5}{\s}} \\
 & \hfil  \textcolor{TABLE_ENTRY_TWO}{\SI{3.2}{\percent}} &  \hfil \textcolor{TABLE_ENTRY_TWO}{\SI{4.5}{\percent}} & \hfil \textcolor{TABLE_ENTRY_TWO}{\SI{4.5}{\percent}} & \hfil \textcolor{TABLE_ENTRY_TWO}{\SI{4.5}{\percent}} & \hfil \textcolor{TABLE_ENTRY_TWO}{\SI{4.5}{\percent}} \\
 & \hfil  \textcolor{TABLE_ENTRY_THREE}{\SI{76}{\percent}} &  \hfil \textcolor{TABLE_ENTRY_THREE}{\SI{53}{\percent}} & \hfil \textcolor{TABLE_ENTRY_THREE}{\SI{13}{\percent}} & \hfil \textcolor{TABLE_ENTRY_THREE}{\SI{13}{\percent}} & \hfil \textcolor{TABLE_ENTRY_THREE}{\SI{11}{\percent}} \\
   & \hfil  \textcolor{TABLE_ENTRY_FOUR}{\SI{57}{\percent}} &  \hfil \textcolor{TABLE_ENTRY_FOUR}{\SI{89}{\percent}} & \hfil \textcolor{TABLE_ENTRY_FOUR}{\SI{205}{\percent}} & \hfil \textcolor{TABLE_ENTRY_FOUR}{\SI{474}{\percent}} & \hfil \textcolor{TABLE_ENTRY_FOUR}{\SI{673}{\percent}} \\
\hspace{0.0002cm} \tabitem Deterministic$\!\!\!\!\!\!\!$ & \hfil  \textcolor{TABLE_ENTRY_ONE}{\SI{3.7}{\s}} &  \hfil  \textcolor{TABLE_ENTRY_ONE}{\SI{11.3}{\s}} & \hfil \textcolor{TABLE_ENTRY_ONE}{\SI{700.6}{\s}} & \hfil \textcolor{TABLE_ENTRY_ONE}{$>$\SI{2000}{\s}} & \hfil \textcolor{TABLE_ENTRY_ONE}{$>$\SI{2000}{\s}} \\
 & \hfil  \textcolor{TABLE_ENTRY_TWO}{\SI{3.2}{\percent}} &  \hfil \textcolor{TABLE_ENTRY_TWO}{\SI{4.8}{\percent}} & \hfil \textcolor{TABLE_ENTRY_TWO}{\SI{4.6}{\percent}} & \hfil \textcolor{TABLE_ENTRY_TWO}{\SI{4.3}{\percent}} & \hfil \textcolor{TABLE_ENTRY_TWO}{\SI{4.5}{\percent}} \\
  & \hfil  \textcolor{TABLE_ENTRY_THREE}{\SI{74}{\percent}} &  \hfil \textcolor{TABLE_ENTRY_THREE}{\SI{67}{\percent}} & \hfil \textcolor{TABLE_ENTRY_THREE}{\SI{0}{\percent}} & \hfil \textcolor{TABLE_ENTRY_THREE}{\SI{0}{\percent}} & \hfil \textcolor{TABLE_ENTRY_THREE}{\SI{0}{\percent}} \\
    & \hfil  \textcolor{TABLE_ENTRY_FOUR}{\SI{53}{\percent}} &  \hfil \textcolor{TABLE_ENTRY_FOUR}{\SI{77}{\percent}} & \hfil \textcolor{TABLE_ENTRY_FOUR}{\SI{266}{\percent}} & \hfil \textcolor{TABLE_ENTRY_FOUR}{\SI{524}{\percent}} & \hfil \textcolor{TABLE_ENTRY_FOUR}{\SI{702}{\percent}} \\
    \bottomrule
    \end{tabular*}
    \vspace{0ex}
\label{table:background_simulation}
\vspace{-0.55cm}
\end{table}

\subsection{Using background simulations to detect collisions}
\label{sec:evaluation_background_simulations}
For each environment, we conduct experiments using random actions and a single background simulation with different time horizons $N$. The results are shown in \mbox{Table \ref{table:background_simulation}}. 
The so-called action adjustment rate indicates the proportion of random actions that are replaced by actions from the backup policy.   
We select the initial states of each environment such that a safe backup trajectory with a time horizon of $N\!=\!30$ exists, eliminating the failure case from Fig. \ref{fig:failure_causes}a.
In the stochastic environments, we also perform experiments under the assumption that the environment behaves deterministically, which additionally rules out the failure case from Fig. \ref{fig:failure_causes}c. 
For $N\!=\!0$, only the current time step is simulated. 
In this case and for $N\!=\!1$, the time until a collision increases only slightly. 
We conclude that a longer time horizon is required to perform an evasive movement with the robot. %
For $N \ge 5$, self-collisions and collisions with the table hardly occur anymore, meaning that moving obstacles are the dominating reason for collisions.
In deterministic environments, collisions with moving obstacles occur less frequently when a longer time horizon is chosen. 
In stochastic environments, however, this is not necessarily the case, as the informative value of a single background simulation decreases over time. 
The higher the time horizon of the background simulation, the greater the computational effort.
In our experiments conducted with an Intel i9-9900K CPU, the computation time exceeded the simulation time for $N \ge 5$. Consequently, real-time execution requires either more computational power or the usage of computationally efficient risk estimators.

Table~\ref{table:braking_trajectories_benchmark} shows a comparison with~\cite{kiemel2022learning}, where braking trajectories are used as backup trajectories.
On average, the braking trajectories require $N=2.9$ time steps to bring the robot joints to a full stop.
Once the robot is stopped, self-collisions and collisions with static obstacles do no longer occur. Consequently, all collisions in the experiments shown in Table~\ref{table:braking_trajectories_benchmark} are caused by moving obstacles.
However, compared to our backup policy trained using model-free RL, the braking trajectories do not evade moving obstacles. 
As a result, the average time to a collision is low when using braking trajectories as backup trajectories.

\begin{figure*}[t]
\vspace{-0.8cm}
\captionsetup[subfigure]{margin=80pt}
    \hspace{-0.18cm}
    \includegraphics[trim=25 0 0 0, clip, width=\textwidth]{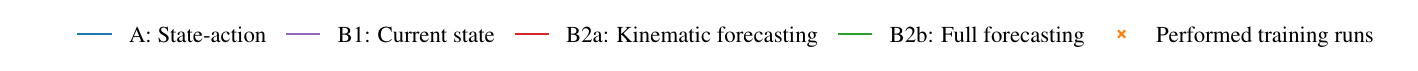}
    \begin{subfigure}[c]{0.33\textwidth}
	    
		\includegraphics[trim=0 0 0 0, clip, height=0.725\textwidth]{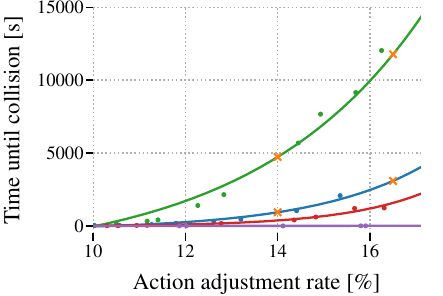}
		
	   \vspace{-0.4cm}\hspace{-0.9cm}\subcaptionbox{Space (deterministic)}[9cm]

	\end{subfigure}
	\hspace{0.025\textwidth}
	\begin{subfigure}[c]{0.33\textwidth}

		\includegraphics[trim=25 0 5 0, clip, height=0.725\textwidth]{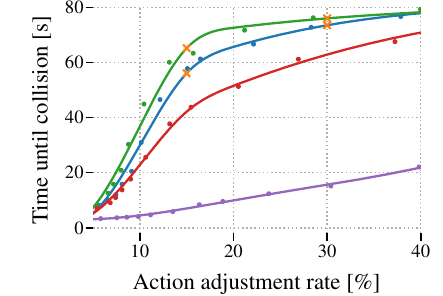}
	    
		\vspace{-0.4cm}\hspace{-1.55cm}\subcaptionbox{Ball (stochastic)}[9cm]

	\end{subfigure} 
	\hspace{-0.014\textwidth}
	\begin{subfigure}[c]{0.33\textwidth}
 
	    \includegraphics[trim=20 0 0 0, clip, height=0.725\textwidth]{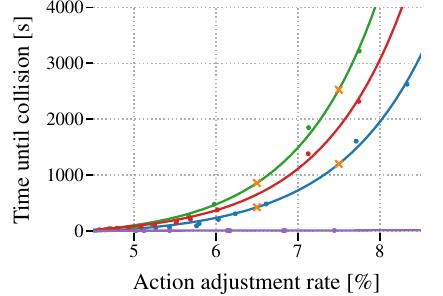}
		
	   \vspace{-0.4cm}\hspace{-1.5cm}\subcaptionbox{Human (stochastic)}[9cm]

	\end{subfigure}
	\vspace{-0.1cm}  
	\caption{%
	The relation between the average time to a collision and the action adjustment rate for a task policy selecting random actions using four different ways to estimate the risk with a risk network as illustrated in Fig. \ref{fig:action_adaptation}. 
	} %
	\label{fig:action_adjustment_rate}
	\vspace{-0.55cm}
\end{figure*}
\begin{table}[h]
     \vspace{-0.155cm}
     \caption{Comparative results using braking trajectories~\cite{kiemel2022learning}.}
     \vspace{-0.2cm}
    \makegapedcells
\begin{tabular*}{0.49\textwidth}{@{}p{16.0mm}p{12mm}p{10.0mm}p{9.0mm}p{19mm}} 
    \toprule
\hspace{0.02cm} Environment & \hfil Time until \hfil  & \multicolumn{1}{c}{$\!$Adjustment} &  \multicolumn{1}{c}{$\!\!\!\!$Time} & \hfil $\!\!\!$Collision with \hfil \\
& \hfil collision \hfil & \multicolumn{1}{c}{$\!$rate}  &  \multicolumn{1}{c}{$\!\!$horizon $N$} \hfil & \hfil $\!\!\!$moving obstacles \hfil \\
    \hline
\hspace{0.0002cm} \tabitem Space  & \hfil \hspace{0.000cm} \SI{13.9}{\s} &  \hspace{0.000cm} \hfil \hspace{0.000cm} \SI{9.7}{\percent} & \hfil $2.9$ & \hfil \SI{100}{\percent} \\
\hspace{0.0002cm} \tabitem Ball & \hfil \hspace{0.000cm}  \SI{21.3}{\s}  & \hspace{0.000cm} \hfil  \hspace{0.000cm} \SI{8.2}{\percent} & \hfil $2.9$ & \hfil \SI{100}{\percent} \\
\hspace{0.0002cm} \tabitem Human & \hfil \hspace{0.000cm} \SI{31.9}{\s} & \hspace{0.000cm} \hfil  \hspace{0.000cm} \SI{7.1}{\percent} & \hfil  $2.9$ & \hfil \SI{100}{\percent} \\
    \bottomrule
    \end{tabular*}
    \vspace{0ex}
\label{table:braking_trajectories_benchmark}
\vspace{-0.65cm}
\end{table}
\begin{table}[bp]
    \vspace{-0.45cm}
    \caption{Training results for a reaching task with \textcolor{TABLE_ENTRY_STATE_ACTION_RISK}{state-action-based~(A)} /  \textcolor{TABLE_ENTRY_STATE_RISK}{state-based~(B2b)} risk estimation.}
    \vspace{-0.15cm}
    \makegapedcells
\begin{tabular*}{0.5\textwidth}{@{}p{20.0mm}p{16mm}p{18mm}p{21mm}} 
    \toprule
    \multicolumn{1}{l}{Initial action}& \multicolumn{1}{c}{Target points}  & \multicolumn{1}{c}{$\!$Time until} & \multicolumn{1}{c}{Final action} \\
    \multicolumn{1}{l}{adjustment rate}& \multicolumn{1}{c}{per second}  & \multicolumn{1}{c}{$\!$collision} & \multicolumn{1}{c}{adjustment rate} \\
    \hline
\hspace{0.12cm} Space & &  \\
\hspace{0.08cm} \tabitem $\,$ \SI{14.0}{\percent}$\!\!\!\!$
    & \hfil $\textcolor{TABLE_ENTRY_STATE_ACTION_RISK}{1.01}$ /  $\textcolor{TABLE_ENTRY_STATE_RISK}{0.94}\,$\hfil  
    & \hfil\textcolor{TABLE_ENTRY_STATE_ACTION_RISK}{\SI{423}{\second}} /  \textcolor{TABLE_ENTRY_STATE_RISK}{\SI{887}{\second}} 
    & \hfil\textcolor{TABLE_ENTRY_STATE_ACTION_RISK}{\SI{9.1}{\percent}} /  \textcolor{TABLE_ENTRY_STATE_RISK}{\SI{8.8}{\percent}}  \\
\hspace{0.08cm} \tabitem $\,$ \SI{16.5}{\percent}$\!\!\!\!$
    & \hfil $\textcolor{TABLE_ENTRY_STATE_ACTION_RISK}{0.93}$ /  $\textcolor{TABLE_ENTRY_STATE_RISK}{0.90}\,$ \hfil  
    &\hfil\textcolor{TABLE_ENTRY_STATE_ACTION_RISK}{\SI{428}{\second}} /  \textcolor{TABLE_ENTRY_STATE_RISK}{\SI{1647}{\second}} 
    & \hfil\textcolor{TABLE_ENTRY_STATE_ACTION_RISK}{\SI{10.0}{\percent}} /  \textcolor{TABLE_ENTRY_STATE_RISK}{\SI{8.8}{\percent}} \\
      
     \hline %
\hspace{0.12cm} Ball  & &  \\  
\hspace{0.08cm} \tabitem $\,$ \SI{15.0}{\percent}$\!\!\!\!$
    & \hfil $\textcolor{TABLE_ENTRY_STATE_ACTION_RISK}{0.70}$ /  $\textcolor{TABLE_ENTRY_STATE_RISK}{0.71}\,$ \hfil  
    &\hfil\textcolor{TABLE_ENTRY_STATE_ACTION_RISK}{\SI{69}{\second}} /  \textcolor{TABLE_ENTRY_STATE_RISK}{\SI{69}{\second}} 
    & \hfil\textcolor{TABLE_ENTRY_STATE_ACTION_RISK}{\SI{18.2}{\percent}} /  \textcolor{TABLE_ENTRY_STATE_RISK}{\SI{17.7}{\percent}} \\
\hspace{0.08cm} \tabitem $\,$ \SI{30.0}{\percent}$\!\!\!\!$
    & \hfil $\textcolor{TABLE_ENTRY_STATE_ACTION_RISK}{0.52}$ /  $\textcolor{TABLE_ENTRY_STATE_RISK}{0.51}\,$ \hfil
    &\hfil\textcolor{TABLE_ENTRY_STATE_ACTION_RISK}{\SI{89}{\second}} /  \textcolor{TABLE_ENTRY_STATE_RISK}{\SI{83}{\second}} 
    & \hfil\textcolor{TABLE_ENTRY_STATE_ACTION_RISK}{\SI{29.4}{\percent}} /  \textcolor{TABLE_ENTRY_STATE_RISK}{\SI{31.1}{\percent}} \\  \hline
\hspace{0.12cm} Human  & &  \\ 
\hspace{0.08cm} \tabitem $\,\:$ \SI{6.5}{\percent}
    & \hfil $\textcolor{TABLE_ENTRY_STATE_ACTION_RISK}{0.79}$ /  $\textcolor{TABLE_ENTRY_STATE_RISK}{0.70}\,$ \hfil  
    &\hfil\textcolor{TABLE_ENTRY_STATE_ACTION_RISK}{\SI{96}{\second}} /  \textcolor{TABLE_ENTRY_STATE_RISK}{\SI{114}{\second}}
    & \hfil\textcolor{TABLE_ENTRY_STATE_ACTION_RISK}{\SI{10.0}{\percent}} /  \textcolor{TABLE_ENTRY_STATE_RISK}{\SI{12.7}{\percent}}  \\
\hspace{0.08cm} \tabitem $\,\:$ \SI{7.5}{\percent}
    & \hfil $\textcolor{TABLE_ENTRY_STATE_ACTION_RISK}{0.49}$ /  $\textcolor{TABLE_ENTRY_STATE_RISK}{0.63}\,$ \hfil  
    &\hfil\textcolor{TABLE_ENTRY_STATE_ACTION_RISK}{\SI{191}{\second}} /  \textcolor{TABLE_ENTRY_STATE_RISK}{\SI{139}{\second}} 
    & \hfil\textcolor{TABLE_ENTRY_STATE_ACTION_RISK}{\SI{16.3}{\percent}} /  \textcolor{TABLE_ENTRY_STATE_RISK}{\SI{16.8}{\percent}} \\
    \bottomrule
    \end{tabular*}
\label{table:training_reaching_task}
\vspace{-0.2cm}
\end{table}

\subsection{Using risk networks to estimate the risk of a collision}
\label{sec:evaluation_risk_networks}
The risk networks used for our evaluation are trained based on previous rollouts of the backup policy with a time horizon of $N=20$.
Once the risk networks are trained, the risk threshold $c_{T\!h}$ can be used to control the average time to a collision. 
With \mbox{$c_{T\!h}=1.0$}, no actions are adjusted by the backup policy.
With \mbox{$c_{T\!h}=0.0$}, all actions are adjusted leading to a rollout of the backup policy (see Table~\ref{table:evaluation_backup_policy}). 
By running experiments with different risk thresholds, the relation between the action adjustment rate and the time to a collision can be determined. 
Based on this relation, it is possible to compare different methods for estimating the risk of an action.
Fig.~\ref{fig:action_adjustment_rate} shows the relation for a task policy selecting random actions when initializing the environments as described in section~\ref{sec:evaluation_background_simulations}.
In all cases, the state-based risk network with full forecasting (B2b) yields the highest time to a collision at a given action adjustment rate.
However, the state-based risk network with full forecasting requires knowledge about the next state of the moving obstacles, which is not needed when using the state-action-based risk network (A).
For our evaluation, we assume that the next state of the moving obstacles can be forecasted.

As shown in Table~\ref{table:training_reaching_task}, we used PPO~\cite{schulman2017proximal} to train task policies for a reaching task using the state-action-based risk estimation (A) and the state-based risk estimation~(B2b). 
The backup policy and the risk networks were used during training and evaluation.
For the reaching task, the task-specific state component $s_{t_{T\!a}}$ encodes the position of a randomly sampled target point and the task policy is rewarded for quickly reaching the target point: 
\begin{align}
    r_{t} = d_{{t}_{T\!a}} - d_{{t+1}_{T\!a}}, \label{eq:reward_reaching_task}
\end{align}
\vspace*{-0.455cm}

\noindent
where $d_{{t}_{T\!a}}$ and $d_{{t+1}_{T\!a}}$ are the distances between the end effector of the robot and the target point at time step $t$ and $t+1$, respectively. 
For each experiment, we used a fixed risk threshold $c_{T\!h}$ so that a specified action adjustment rate was obtained at the beginning of the training process. %
In Table~\ref{table:training_reaching_task}, we additionally indicate the final action adjustment rate obtained after training. While the resulting task performance and the average time to a collision depend on the task policy learned during the training process, we identified the following tendencies in our experiments: Selecting a higher initial action adjustment rate also increased the average time to a collision for trained agents.
On the other hand, the training performance, measured by the number of target points reached per second, decreased when selecting a higher initial action adjustment rate.
In the space environment, the state-based risk estimation led to significantly fewer collisions than the state-action-based risk estimation. In the other two environments, both risk estimation methods led to similar training results. 

\vspace{-0.1cm}
\subsection{Benchmarking with a QP-based method}
We compare our approach with a QP-based method~\cite{pham2018optlayer} used during training and evaluation.
To keep the number of QP constraints low, we simplify our space environment by removing the table and ignoring self-collisions.
\begin{table}[t]
     \caption{QP baseline in a simplified space environment.}
     \vspace{-0.15cm}
    \makegapedcells
\begin{tabular*}{0.49\textwidth}{@{}p{18mm}p{13mm}p{12.0mm}p{14.0mm}p{12mm}} 
    \toprule
\multirow[t]{4}{*}{\hspace{0.02cm}}
 & \multicolumn{2}{c}{Untrained agent} & \multicolumn{2}{c}{Trained agent}  \\
& \multicolumn{1}{c}{Adjustment}  & \multicolumn{1}{c|}{Time until} &  \multicolumn{1}{c}{Target points}  & \multicolumn{1}{c}{Time until}  \\
& \multicolumn{1}{c}{rate}  & \multicolumn{1}{c|}{collision} &  \multicolumn{1}{c}{per second}  & \multicolumn{1}{c}{collision}  \\
    \hline
\hspace{-0.02cm}\tabitem QP method  & \hfil $\;$\SI{12.3}{\percent} &  \hfil \SI{35.6}{\second} & \hfil $0.70$ & \hfil \SI{2216.6}{\s} \\
\hspace{-0.02cm}\tabitem State-action (A)$\!\!\!\!\!\!\!\!\!\!\!$ &  \hfil $\;$\SI{7.1}{\percent} &  \hfil $\;$\SI{855.2}{\second} & \hfil $0.82$ & \hfil \SI{4325.2}{\s} \\
    \bottomrule
    \end{tabular*}
    \vspace{-0.55cm}
\label{table:qp_benchmark}
\end{table}
In addition, we model the moving obstacles and the end effector of the robot as a sphere. Table~\ref{table:qp_benchmark} shows the results of the comparison. 
It can be seen, that our method leads to fewer collisions while adjusting fewer actions. Moreover, the trained agent learns to reach a larger number of target points per time. 

\vspace{-0.23cm}
\subsection{Learning a basketball task}
\label{sec:basketball_task}
In addition to the reaching task, we use the same backup networks and the state-action-based risk networks to learn a basketball task, where the robot is rewarded for placing balls into basketball hoops moving around the robot. 
To this end, the task-specific action component $a_{t_{T\!a}}$ controls the speed at which a ball should leave a tube attached to the robot.
As shown in Table~\ref{table:basketball} and in our video, the robot manages to learn the task while keeping the risk of a collision low. 

\begin{table}[h]
     \vspace{-0.1cm}
     \caption{Training results for a basketball task.}
     \vspace{-0.15cm}
    \makegapedcells
\begin{tabular*}{0.49\textwidth}{@{}p{14mm}p{14mm}p{12.0mm}p{14.0mm}p{12mm}} 
    \toprule
\multirow[t]{3}{*}{\hspace{0.02cm}}
 & \multicolumn{2}{c}{Untrained agent} & \multicolumn{2}{c}{Trained agent}  \\
& \multicolumn{1}{c}{Hoops scored}  & \multicolumn{1}{c|}{Time until} &  \multicolumn{1}{c}{Hoops scored}  & \multicolumn{1}{c}{Time until}  \\
& \multicolumn{1}{c}{per second}  & \multicolumn{1}{c|}{collision} &  \multicolumn{1}{c}{per second}  & \multicolumn{1}{c}{collision}  \\
    \hline
\hspace{0.0002cm} \tabitem Space  & \hfil $0.007$ &  \hfil \SI{831.5}{\second} & \hfil $1.84$ & $\!\!>\SI{20000}{\s}$ \\
\hspace{0.0002cm} \tabitem Ball & \hfil $0.010$ &  \hfil \SI{78.4}{\second} & \hfil $1.38$ & \hfil \SI{86.2}{\s} \\
\hspace{0.0002cm} \tabitem Human & \hfil $0.008$ &  \hfil \SI{1258.1}{\second} & \hfil $1.19$ & \hfil $\!\!>\SI{20000}{\s}$ \\
    \bottomrule
    \end{tabular*}
    \vspace{0ex}
\label{table:basketball}
\vspace{-0.6cm}
\end{table}

\subsection{Sim-to-real transfer and real-time capability}
\label{sec:sim-to-real}

Our method produces jerk-limited trajectories that can be tracked by a real robot without overloading the robot joints. 
In addition, the task policy, the backup policy and the risk estimation require only a small amount of computing power as they are based on neural networks (see Table~\ref{table:computation_times}). %
In our video, we demonstrate a successful sim-to-real transfer of a reaching policy in the space environment using a KUKU iiwa robot. %
For the experiment, the space station and the asteroid are assumed to move as during the \mbox{training phase}.

\begin{table}[h]
    \vspace{-0.1cm}
     \caption{Computation time per simulation time.}
     \vspace{-0.15cm}
    \makegapedcells
\begin{tabular*}{0.49\textwidth}{@{}p{21mm}p{17mm}p{11.0mm}p{11.0mm}p{11mm}} 
    \toprule
\multicolumn{1}{l}{Risk network}
 & \multicolumn{1}{c}{State-action} & \multicolumn{3}{c}{State} \\
& \multicolumn{1}{c|}{A}  & \multicolumn{1}{c}{B1} &  \multicolumn{1}{c}{B2a}  & \multicolumn{1}{c}{B2b} \\
    \hline
\hspace{0.0002cm} \tabitem Space  & \hfil \SI{8.7}{\percent} &   \hfil \SI{8.6}{\percent} &  \hfil \SI{8.6}{\percent} &  \hfil \SI{8.8}{\percent} \\
\hspace{0.0002cm} \tabitem Ball &  \hfil \SI{10.7}{\percent} &   \hfil \SI{10.5}{\percent} &  \hfil \SI{11.1}{\percent} &  \hfil \SI{10.9}{\percent} \\
\hspace{0.0002cm} \tabitem Human & \hfil \SI{18.5}{\percent} &   \hfil \SI{17.6}{\percent} &  \hfil \SI{18.2}{\percent} &  \hfil \SI{18.6}{\percent} \\
    \bottomrule
    \end{tabular*}
    \vspace{0ex}
\label{table:computation_times}
\vspace{-0.55cm}
\end{table}

\section{Conclusion and future work}
\label{sec:conclusion}
We presented an approach to learn robot trajectories in the presence of moving obstacles while keeping the risk of collisions low. 
We confirmed the effectiveness of our approach by successfully learning a reaching task and a basketball task in three different environments %
and demonstrated real-time capability by running a policy trained in simulation on a real robot. 
An interesting direction for future research is to investigate measures to reduce the action adjustments caused by the backup policy, e.g., by searching for a safe action close to the desired action of the task policy.
It would also be interesting to investigate the impact of uncertainty in sensor measurements as an additional source of stochasticity.
Furthermore, we are interested in applying our approach to other robotic systems and other safety constraints, for instance by learning to control a bipedal robot without falling over.

\vspace*{-0.2cm}

\bibliographystyle{IEEEtran}
\bibliography{root}

\begin{thebibliography}{10}
\providecommand{\url}[1]{#1}
\csname url@rmstyle\endcsname
\providecommand{\newblock}{\relax}
\providecommand{\bibinfo}[2]{#2}
\providecommand\BIBentrySTDinterwordspacing{\spaceskip=0pt\relax}
\providecommand\BIBentryALTinterwordstretchfactor{4}
\providecommand\BIBentryALTinterwordspacing{\spaceskip=\fontdimen2\font plus
\BIBentryALTinterwordstretchfactor\fontdimen3\font minus \fontdimen4\font\relax}
\providecommand\BIBforeignlanguage[2]{{%
\expandafter\ifx\csname l@#1\endcsname\relax
\typeout{** WARNING: IEEEtran.bst: No hyphenation pattern has been}%
\typeout{** loaded for the language `#1'. Using the pattern for}%
\typeout{** the default language instead.}%
\else
\language=\csname l@#1\endcsname
\fi
#2}}

\bibitem{pham2018optlayer}
T.-H. Pham, G.~De~Magistris, and R.~Tachibana, ``Optlayer-practical constrained optimization for deep reinforcement learning in the real world,'' in \emph{2018 IEEE International Conference on Robotics and Automation (ICRA)}.\hskip 1em plus 0.5em minus 0.4em\relax IEEE, 2018, pp. 6236--6243.

\bibitem{faverjon1987local}
B.~Faverjon and P.~Tournassoud, ``A local based approach for path planning of manipulators with a high number of degrees of freedom,'' in \emph{Proceedings. 1987 IEEE international conference on robotics and automation}, vol.~4.\hskip 1em plus 0.5em minus 0.4em\relax IEEE, 1987, pp. 1152--1159.

\bibitem{fisac2018general}
J.~F. Fisac, A.~K. Akametalu, M.~N. Zeilinger, S.~Kaynama, J.~Gillula, and C.~J. Tomlin, ``A general safety framework for learning-based control in uncertain robotic systems,'' \emph{IEEE Transactions on Automatic Control}, vol.~64, no.~7, pp. 2737--2752, 2018.

\bibitem{wabersich2021predictive}
K.~P. Wabersich and M.~N. Zeilinger, ``A predictive safety filter for learning-based control of constrained nonlinear dynamical systems,'' \emph{Automatica}, vol. 129, p. 109597, 2021.

\bibitem{kiemel2022learning}
J.~C. Kiemel and T.~Kr{\"o}ger, ``Learning collision-free and torque-limited robot trajectories based on alternative safe behaviors,'' in \emph{2022 IEEE-RAS 21st International Conference on Humanoid Robots (Humanoids)}.\hskip 1em plus 0.5em minus 0.4em\relax IEEE, 2022, pp. 223--230.

\bibitem{yang2022safe}
T.-Y. Yang, T.~Zhang, L.~Luu, S.~Ha, J.~Tan, and W.~Yu, ``Safe reinforcement learning for legged locomotion,'' in \emph{2022 IEEE/RSJ International Conference on Intelligent Robots and Systems (IROS)}.\hskip 1em plus 0.5em minus 0.4em\relax IEEE, 2022, pp. 2454--2461.

\bibitem{thananjeyan2021recovery}
B.~Thananjeyan, A.~Balakrishna, S.~Nair, M.~Luo, K.~Srinivasan, M.~Hwang, J.~E. Gonzalez, J.~Ibarz, C.~Finn, and K.~Goldberg, ``Recovery rl: Safe reinforcement learning with learned recovery zones,'' \emph{IEEE Robotics and Automation Letters}, vol.~6, no.~3, pp. 4915--4922, 2021.

\bibitem{altman1999constrained}
E.~Altman, \emph{Constrained Markov decision processes}.\hskip 1em plus 0.5em minus 0.4em\relax CRC Press, 1999.

\bibitem{achiam2017constrained}
J.~Achiam, D.~Held, A.~Tamar, and P.~Abbeel, ``Constrained policy optimization,'' in \emph{International Conference on Machine Learning}, 2017.

\bibitem{liu2020ipo}
Y.~Liu, J.~Ding, and X.~Liu, ``Ipo: Interior-point policy optimization under constraints,'' in \emph{Proceedings of the AAAI Conference on Artificial Intelligence}, vol.~34, no.~04, 2020, pp. 4940--4947.

\bibitem{brunke2022safe}
L.~Brunke, M.~Greeff, A.~W. Hall, Z.~Yuan, S.~Zhou, J.~Panerati, and A.~P. Schoellig, ``Safe learning in robotics: From learning-based control to safe reinforcement learning,'' \emph{Annual Review of Control, Robotics, and Autonomous Systems}, vol.~5, pp. 411--444, 2022.

\bibitem{zhao2021model}
W.~Zhao, T.~He, and C.~Liu, ``Model-free safe control for zero-violation reinforcement learning,'' in \emph{5th Annual Conference on Robot Learning}, 2021.

\bibitem{berkenkamp2017safe}
F.~Berkenkamp, M.~Turchetta, A.~Schoellig, and A.~Krause, ``Safe model-based reinforcement learning with stability guarantees,'' \emph{Advances in neural information processing systems}, vol.~30, 2017.

\bibitem{koller2018learning}
T.~Koller, F.~Berkenkamp, M.~Turchetta, and A.~Krause, ``Learning-based model predictive control for safe exploration,'' in \emph{2018 IEEE conference on decision and control (CDC)}.\hskip 1em plus 0.5em minus 0.4em\relax IEEE, 2018, pp. 6059--6066.

\bibitem{cheng2019end}
R.~Cheng, G.~Orosz, R.~M. Murray, and J.~W. Burdick, ``End-to-end safe reinforcement learning through barrier functions for safety-critical continuous control tasks,'' in \emph{Proceedings of the AAAI Conference on Artificial Intelligence}, vol.~33, no.~01, 2019, pp. 3387--3395.

\bibitem{wang2023safe}
K.~Wang, C.~Mu, Z.~Ni, and D.~Liu, ``Safe reinforcement learning and adaptive optimal control with applications to obstacle avoidance problem,'' \emph{IEEE Transactions on Automation Science and Engineering}, vol.~21, no.~3, pp. 4599--4612, 2024.

\bibitem{hans2008safe}
A.~Hans, D.~Schneega{\ss}, A.~M. Sch{\"a}fer, and S.~Udluft, ``Safe exploration for reinforcement learning.'' in \emph{ESANN}, 2008, pp. 143--148.

\bibitem{ShortPaperMotionSafety}
T.~Fraichard, ``A short paper about motion safety,'' in \emph{Proceedings 2007 IEEE International Conference on Robotics and Automation}.\hskip 1em plus 0.5em minus 0.4em\relax IEEE, 2007, pp. 1140--1145.

\bibitem{rubrecht2012motion}
S.~Rubrecht, V.~Padois, P.~Bidaud, M.~De~Broissia, and M.~D.~S. Simoes, ``Motion safety and constraints compatibility for multibody robots,'' \emph{Autonomous Robots}, vol.~32, no.~3, pp. 333--349, 2012.

\bibitem{srinivasan2020learning}
K.~Srinivasan, B.~Eysenbach, S.~Ha, J.~Tan, and C.~Finn, ``Learning to be safe: Deep rl with a safety critic,'' \emph{arXiv:2010.14603}, 2020.

\bibitem{kiemel2021learning}
J.~C. Kiemel and T.~Kr{\"o}ger, ``Learning robot trajectories subject to kinematic joint constraints,'' in \emph{2021 IEEE International Conference on Robotics and Automation (ICRA)}.\hskip 1em plus 0.5em minus 0.4em\relax IEEE, 2021, pp. 4799--4805.

\bibitem{schulman2017proximal}
J.~Schulman, F.~Wolski, P.~Dhariwal, A.~Radford, and O.~Klimov, ``Proximal policy optimization algorithms,'' \emph{arXiv:1707.06347}, 2017.

\end{thebibliography}

\end{document}